%% file: Athindran_short.tex
\documentclass[sigconf]{acmart}
\usepackage[utf8]{inputenc} % allow utf-8 input
\usepackage[T1]{fontenc}    % use 8-bit T1 fonts
\usepackage{hyperref}       % hyperlinks
\usepackage{url}            % simple URL typesetting
\usepackage{booktabs}       % professional-quality tables
\usepackage{amsfonts}       % blackboard math symbols
\usepackage{nicefrac}       % compact symbols for 1/2, etc.
\usepackage{microtype}      % microtypography
\usepackage{lipsum}
\usepackage{graphicx}
\usepackage{subfig}
\usepackage{multirow}
\usepackage{tikz}
\usetikzlibrary{shapes,calc,arrows,chains,positioning,shadows,trees}
\usepackage{booktabs} % For formal tables

\copyrightyear{2019} 
\acmYear{2019} 
\setcopyright{acmlicensed}
\acmConference[CoDS-COMAD '19]{6th ACM IKDD CoDS and 24th COMAD}{January 3--5, 2019}{Kolkata, India}
\acmBooktitle{6th ACM IKDD CoDS and 24th COMAD (CoDS-COMAD '19), January 3--5, 2019, Kolkata, India}
\acmPrice{15.00}
\acmDOI{10.1145/3297001.3297020}
\acmISBN{978-1-4503-6207-8/19/01}
\acmConference[CODS-COMAD'19]{The ACM India Joint International Conference on Data Science and Management of Data}{January 2019} {Kolkata, India}
\acmYear{2019}
\copyrightyear{2019}

\begin{document}
\title{Pack and Detect: Fast Object Detection in Videos Using Region-of-Interest Packing}

\author{Athindran Ramesh Kumar}
\affiliation{
  \institution{Indian Institute of Technology Madras}
  \streetaddress{Department of Computer Science}
%  \city{Chennai}
%  \state{Tamil Nadu}
%  \country{India} 
}
\email{r.athindran@gmail.com}

\author{Balaraman Ravindran}
\affiliation{
  \institution{Indian Institute of Technology Madras}
  \streetaddress{Robert Bosch Center for Data Science and Artificial Intelligence}
%  \city{Chennai}
%  \state{Tamil Nadu}
%  \country{India}
}
\email{ravi@cse.iitm.ac.in}

\author{Anand Raghunathan}
\affiliation{
  \institution{Purdue University}
  \streetaddress{Department of Electrical Engineering}
%  \city{West Lafayette}
%  \state{Indiana}  
}
\email{raghunathan@purdue.edu}

\renewcommand{\shortauthors}{R. Athindran et al.}

\begin{abstract}
Object detection in videos is an important task in computer vision for
various applications such as object tracking, video summarization and
video search. Although great progress has been made in improving the
accuracy of object detection in recent years due to the rise of deep
neural networks, the state-of-the-art algorithms are highly
computationally intensive. In order to address this challenge, we make
two important observations in the context of videos: (i) Objects often
occupy only a small fraction of the area in each video frame, and (ii)
There is a high likelihood of strong temporal correlation between
consecutive frames. Based on these observations, we propose Pack and
Detect (PaD), an approach to reduce the computational requirements of
object detection in videos. In PaD, only selected video frames called
anchor frames are processed at full size.  In the frames that lie
between anchor frames (inter-anchor frames), regions of interest
(ROIs) are identified based on the detections in the previous
frame. We propose an algorithm to pack the ROIs of each inter-anchor
frame together into a reduced-size frame. The computational
requirements of the detector are reduced due to the lower size of the
input. In order to maintain the accuracy of object detection, the
proposed algorithm expands the ROIs greedily to provide additional
background around each object to the detector. PaD can use any
underlying neural network architecture to process the full-size and
reduced-size frames. Experiments using the ImageNet video object
detection dataset indicate that PaD can potentially reduce the number
of FLOPS required for a frame by $4\times$. This leads to an overall
increase in throughput of $1.25\times$ on a 2.1 GHz Intel Xeon server
with a NVIDIA Titan X GPU at the cost of $1.1\%$ drop in accuracy.
\end{abstract}


\maketitle

\input{Athindran_body_short}

\bibliographystyle{ACM-Reference-Format}
\bibliography{references}

\end{document}

%% file: Athindran_body_short.tex
\section{Introduction}
The task of object detection in videos
\cite{KangTCNN16,zhu2017deep,Kang_2016_CVPR,Kang2017object,zhu2017flow,Zhu_2018_CVPR,TSSD,pan2018recurrent,mobilevideo,fastYOLO}
has been gaining attention in recent years. It serves as an
important preprocessing task for object tracking and for
several other video processing tasks such as video summarization and
video search.
%Object detection and tracking can aid various
%important applications such as traffic monitoring, pedestrian
%tracking, animal monitoring and environment survey using drones.
Many object detection applications require video frames to be
processed in real-time in resource-constrained environments. It is
thus imperative to design systems that can detect objects in videos
accurately, but also in a computationally efficient manner.

\par Still-image object detection has been studied extensively in the
past. The accuracy and speed of still-image object detection have
improved by leaps and bounds in recent years due to advances in deep
convolutional neural networks (CNN). Recent CNN-based
object detectors include Faster-RCNN \cite{NIPS2015_FRCNN}, SSD
\cite{liu2016ssd}, YOLO \cite{CVPR_YOLO,yolo9000,yolov3} and
RFCN\cite{dai2016r}. These still-image object detectors can be
extended for object detection in videos by using them on a per-frame
basis. However, this is inefficient as there is a strong temporal
correlation between frames in a video. This temporal redundancy can be
leveraged either to improve the accuracy or speed of object
detection. In the recent past, there have been several attempts to
improve the accuracy of object detection in videos by either
integrating the bounding boxes
\cite{KangTCNN16,Kang_2016_CVPR,han2016seq,penglinking} or features
\cite{Kang2017object,zhu2017flow,Zhu_2018_CVPR,hetangimpression}
across frames. However, there has not been enough attention on
leveraging this temporal redundancy to improve speed. Some exceptions
to this norm are
\cite{zhu2017deep,TSSD,pan2018recurrent,mobilevideo,fastYOLO}. In this
work, we propose a method, Pack and Detect (PaD), for fast object
detection in videos that can work with any underlying object detector.
%\par
%Neural networks are in general very computationally intensive. For example, processing a video at $300 \times 300$ resolution using the SSD300 (Single Shot Detector) object detection network with VGG16 as backbone at 30 fps requires 1.87 trillion FLOPS/s. Several attempts \cite{szesurvey} have been made to make them less compute-hungry. However, firstly, all of these methods do not exploit the specific opportunities that can be leveraged in the context of videos. Secondly, all of these methods try to modify the network and do not try to compress the inputs to be fed into the network.

\par PaD leverages two key opportunities in the context of object
detection in videos. First, the objects of interest often occupy only
a small fraction of an image. Second, there is a strong correlation
between successive frames in a video. In PaD, only selected frames
called anchor frames are passed in their entirety to the underlying
object detector. In frames that lie between anchor frames
(inter-anchor frames), the detections from the previous frame are used
to identify ROIs in the image where an object could potentially be
located. The ROIs are packed in a reduced-size image that is fed into
the detector, resulting in lower computational requirements.

\par We propose a ROI packing algorithm based on the
following criteria:
\begin{enumerate}
\item Each ROI is expanded to provide as much background context as possible to maintain the accuracy of the detector.
\item There is minimal loss of resolution and no change in aspect ratio to maintain the accuracy of the detector.
\item Each object is present in a unique ROI. 
\item The space in the reduced-size frame is used as efficiently as possible.
\end{enumerate}
\par

%Although we focus on two frame sizes, this approach can be generalized into
%larger numbers of sizes.
We evaluate PaD by implementing it on top of the SSD300 object
detector and evaluating it with the ImageNet video object detection
dataset. Our results indicate that PaD reduces the FLOP count for
reduced-size frames by around $4\times$. Overall, PaD achieves
$1.25\times$ increase in throughput with only a $1.1\%$ drop in
accuracy.
%We present results that provide insight into the inner
%workings of the algorithm.

\section{Related Work}
\label{sec:rel_work}
\subsection*{Object Detection in Videos}
The temporal redundancy present in videos has been exploited before to
improve the accuracy and speed of object detection. In
\cite{KangTCNN16,Kang_2016_CVPR,han2016seq,penglinking}, the
aggregation of information from neighbouring frames is done at the
bounding box level to improve accuracy. In \cite{KangTCNN16},
per-frame object detection is combined with multi-context suppression,
motion-guided propagation and object tracking to improve detection
accuracy. In \cite{han2016seq,penglinking}, non-maximum suppression is
done over bags of frames. In
\cite{Kang2017object,zhu2017flow,Zhu_2018_CVPR,hetangimpression},
integration of the CNN features across neighbouring frames is used to
improve accuracy. In \cite{Kang2017object}, a CNN is combined with a
Long Short Term Memory (LSTM) to obtain temporal features for object
detection. In \cite{zhu2017flow, Zhu_2018_CVPR,hetangimpression}, the
features from neighbouring frames are aggregated together using
optical flow information to improve feature quality. These
methods~\cite{KangTCNN16,Kang_2016_CVPR,han2016seq,penglinking,Kang2017object}
pose large computation requirements, making them often unsuitable for
real-time processing.

\par On the other hand,
\cite{zhu2017deep,TSSD,pan2018recurrent,mobilevideo,fastYOLO} are
relatively faster methods aimed at object detection in videos. The
methods in \cite{TSSD,mobilevideo,fastYOLO} are faster by virtue of
using a faster still-image object detector or an efficient backbone
network. In \cite{zhu2017deep}, the feature maps from selected anchor
frames are transferred to neighbouring frames by warping them with
optical flow information, leading to reduced computation. In
\cite{pan2018recurrent}, neighbouring frames are subtracted to give
rise to a sparse input that is processed with a sparsity-aware
hardware accelerator~\cite{Han2016} to achieve computational
savings.

\par PaD differs vastly from the prior methods proposed to speed-up
video object detection. PaD can be used alongside previous methods
such as \cite{zhu2017deep} and on top of existing efficient object
detectors that operate on a per-frame
basis~\cite{TSSD,mobilevideo,fastYOLO}. Moreover, PaD does not require
any specialized hardware accelerator like in \cite{pan2018recurrent}
to obtain computational savings.

\subsection*{Efficient Neural Networks}
Several efforts have attempted to reduce the computational
requirements of neural networks. Quantization with retraining was
shown to improve the efficiency of neural network implementations
in~\cite{axnn}. Deep compression~\cite{HanMD15} combined pruning,
trained quantization and weight compression and demonstrated large
speedups on a custom hardware accelator~\cite{Han2016}. Subsequent
efforts have explored structured sparsity~\cite{NIPS2016_ssparsity} by
pruning
filters~\cite{thinet,NIPS2016_ssparsity,Guo2017,deepakrandom,haopruning}
of a CNN. MobileNet~\cite{mobilenet} replaces the standard convolution
with a combination of depth-wise and point-wise convolution to reduce
computation. SqueezeNet~\cite{squeezenet} uses network architecture
modifications to reduce the number of computations and
memory. Scalable-effort classifiers reduce computational requirements
by first using lower-complexity classifiers to process an input and
subsequently using higher-accuracy classifiers only when
needed~\cite{venkataramani2015scalable}. A similar approach is taken
by Big-Little networks~\cite{biglittle}.
%\par Most of the methods for achieving computational efficiency in
%neural networks are static methods to be employed at design
%time. However, there are some exceptions.
Conditional computation
\cite{bengiocondcomp} selectively activates certain parts of the
network depending on the input. The policy for deciding which parts of
the network to activate is learnt using reinforcement
learning. Dynamic deep neural networks (D2NN) \cite{LiuD17} work in
a similar manner to conditional computation and turn on/off regions of
the network using reinforcement learning. DyVEDeep
\cite{GanapathyVRR17} reduces computations in neural networks
dynamically by using three strategies - saturation prediction and
early termination, significance driven selective sampling and
similarity-based feature map approximation.

\par The above methods focus on modifications to the network to reduce
computations and achieve speedup. In this work, we take a
complementary approach and compress the inputs that we feed into the
network. Hence, PaD is orthogonal to most existing techniques and can
be used in combination with them.

\subsection*{Visual Attention Mechanism}
Inspired by human vision, there have been several attempts
\cite{NIPS2010_4089,Ranzato14,Mnih2014,ba2014multiple,kosiorek2017hierarchical,kahou2015ratm}
to reduce computation by processing an image as a sequence of glimpses
rather than as a whole. The notion of a foveal glimpse is somewhat
similar to the idea of ROI discussed here. However, there are several
important differences. A foveal glimpse is a high resolution crop of
an important region in the image that is crucial to the task at
hand. In our work, we pack all the ROIs together in a single frame and
do not process them sequentially. Further, the location of ROIs is
inferred from the detections in the previous frame in a video and does
not need an attention mechanism. Also, a foveal glimpse obtains crops
by extracting pixels close to the location target at high resolution
and pixels far from the location target at low resolution. We do not
employ multi-resolution processing. Hence, our work, although inspired
from the notion of foveal attention is considerably different.

\subsection*{Multiple Object Tracking}
Object detection in video is a precursor to the problem of multiple
object tracking. Once the objects are detected in the video, the
detections are linked together to form a track. This problem is
studied separately from the object detection problem in the
literature. The ImageNet VID dataset used in this work does not have
ground truth labels to measure the tracking metrics. In the MOT
challenge \cite{milan2016mot16}, the detections that are input to the
tracker are provided with the dataset. Several popular trackers such
as
\cite{leal2016learning,xiang2015learning,milan2017online,sadeghian2017tracking}
have garnered attention through the challenge. While the use of
detection and tracking to complement each other to improve accuracy or
speed is possible, it is not well studied in the literature. In
\cite{feichtenhofer2017detect}, the detection and tracking have been
used in a complementary fashion to improve the accuracy. In future
work, we will explore the potential of combining detection and
tracking to improve speed.

\section{Motivation}
\label{sec:motivation}
\subsection{Occupancy of objects in frames}
\begin{figure}
  \centering
  \includegraphics[scale=0.4]{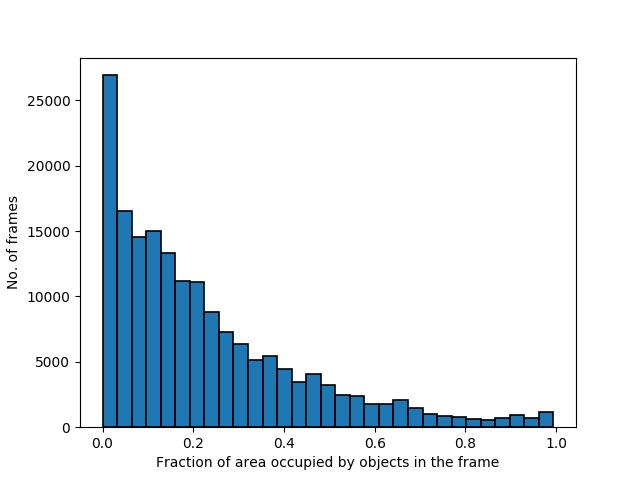}
  \caption{Histogram of object occupancy ratio}
  \label{fig:occupancy}
\end{figure}

PaD leverages the hypothesis that the objects of interest occupy only
a small fraction of the area in the frame. We support this hypothesis
using statistics from a popular video dataset. Figure
\ref{fig:occupancy} is a histogram of the object occupancy ratio
in the ImageNet VID validation set containing $555$ videos
with $176126$ frames. From the figure, we see that the objects occupy
only $22.7\%$ of the frame on average. In a vast majority of the
frames, the objects occupy less than $30\%$ of the frame.

\subsection{Temporal correlation of object locations across frames}
\begin{figure}
  \centering
  \includegraphics[scale=0.4]{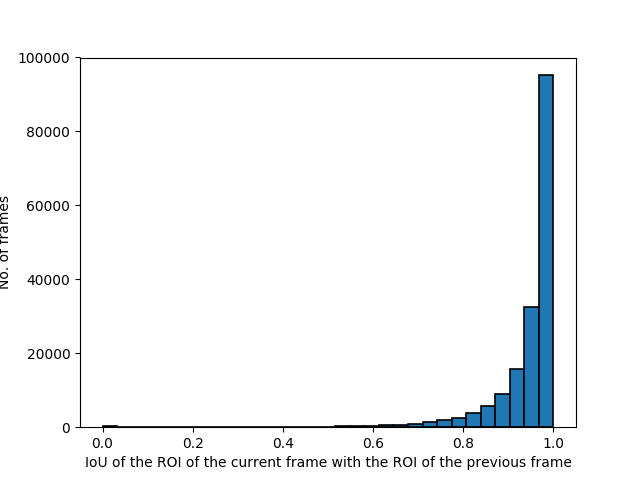}
  \caption{Histogram of Intersection over Union (IoU) of regions
    containing objects between consecutive frames}
  \label{fig:ROIoverlap}
\end{figure}

It is well known that successive frames in a video are likely to be
highly correlated. We illustrate this through a
statistical analysis of the ImageNet VID validation set.  Figure
\ref{fig:ROIoverlap} presents a histogram of the object occupancy area
Intersection over Union (IoU) statistics between consecutive frames in
the dataset. In the figure we can clearly see a sharp peak close to
$1$. On average, the IoU of areas containing objects between consecutive
frames is $94.4\%$.

\section{Pack-and-Detect: Approach and Algorithms}
\label{sec:Methodology}
\subsection{Overview}
An overview of the PaD approach is presented in Figure
\ref{fig:arch}. Full-sized video frames are processed at regular
intervals (by designating the first of every $d$ frames as an anchor
frame). In other frames, ROIs are identified based on the
locations of the detections from the previous frame. Only detections
with a minimum confidence threshold $\tau$ are taken into
consideration. An ROI packing algorithm attempts to pack the ROIs into a
reduced-size frame. If the packing is successful, then the
reduced-size frame is processed instead, giving rise to computational
savings. Once the reduced-size frame is processed using the CNN
detector, the object locations are mapped back to the original frame.
However, if the packing is not successful, then the frame is processed
at full size, incurring an overhead due to checking whether ROI
packing is possible. We demonstrate that this tradeoff is often
favorable, resulting in a net improvement in the speed of object
detection.

\begin{figure}
\centering
\begin{tikzpicture}[auto, node distance=3cm and 10 cm,>=latex]
\tikzset{
  base/.style={draw, on grid,align=center, minimum height=8ex},
  proc/.style={base, rectangle, text width=6em},
  test/.style={base, diamond, text width=5em},
  coord/.style={coordinate, on grid},
}
\node[test,xshift=10mm](p0){\small Anchor frame?};
\node[proc,below of = p0](p1){\small Process frame at full size};
\node[proc,above of = p0](p2){\small Identify ROIs in frame};
\node[proc,right of = p2, xshift = 20mm](p3){\small Run ROI packing algorithm};
\node[test, below of = p3](p4){\small ROI packing successful?};
\node[proc, below of = p4](p5){\small Process frame at lower size};
\path (p0.south) to node [near start, xshift=0.2em] {$Yes$} (p1);
\draw [*->] (p0.south) -- (p1);
\path (p0.north) to node [near start, xshift=0.2em] {$No$} (p2);
\draw [*->] (p0.north) -- (p2);
\path (p2.east) to node [near start, xshift=0.2em] {} (p3);
\draw [->] (p2.east) -- (p3);
\path (p3.south) to node [near start, xshift=0.2em] {} (p4);
\draw [->] (p3.south) -- (p4);
\path (p4.south) to node [near start, xshift=0.2em] {$Yes$} (p5);
\draw [*->] (p4.south) -- (p5);
\path (p4.west) to node [near start, xshift=0.2em] {$No$} (p1);
\draw [*->] (p4.west) -- (p1);
%\node [coord, above of = p4]  (c1)  {};
%\node [coord, left of = p1, node distance = 6em] (c2){};
%\path (p4.north) to node [xshift=2em] {$No$} (c1);
%  \draw [*-] (p4.north) -- (c1) -| (c2);
%\path (c2.east) to node [xshift=1em]{}(p1);
%  \draw [->] (c2.east) -- (p1);   
\end{tikzpicture}
\caption{Overall approach of PaD} \label{fig:arch}
\end{figure}

\subsection{ROI packing algorithm}
Figure \ref{fig:flow1} describes the ROI packing algorithm. As a first
step in the algorithm, we construct a graph where nodes represent ROIs
and an edge connects two nodes if the corresponding ROIs intersect. We
find all connected components of this graph. We then find the
enclosing bounding box over the union of ROIs in each connected
component. We iterate the connected components algorithm until the
final bounding boxes do not overlap. This constraint is important
because if two bounding boxes overlap, then parts of the same object
could be present two or more times in the packed frame. Once the
number and size of the bounding boxes are decided, the layout of the
bounding boxes is determined by using the algorithm presented in
Figure \ref{fig:tiling}. Once the layout is decided, a check is done
to see whether the bounding boxes can fit in the layout. If it is not
possible to fit the bounding boxes in the layout, the image is
processed at full size.

\par If the bounding boxes can be fit in the layout, a post-processing
step is performed as described below. Our experiments indicated that
neural network based object detectors are often overfit to the
background context of the object to be detected. Consequently, the
accuracy of the object detector degrades if there is no background
context. To address this challenge, we extend each bounding box to
provide as much context as possible to the detector. The algorithm
for extending the bounding boxes works as follows. We decide
whether to first extend the boxes horizontally or
vertically. For the sake of discussion, let us assume that the choice
is to first extend all the bounding boxes horizontally. We find all
the bounding boxes that could potentially intersect when extended
horizontally. We extend all bounding boxes horizontally until the
layout size is reached or the bounding boxes start intersecting with
each other. Then, we repeat the same procedure in the other
dimension. Once the final bounding boxes are decided, the
corresponding regions in the image are extracted and the reduced-size
frame is composed according to the determined layout.
%\par
%There are a lot of heuristic rules in the ROI packing algorithm. All of these rules have been decided after thorough analysis of the setup using experiments. In section \ref{sec:results}, we show some results illustrating that this sophisticated ROI packing algorithm is superior to a naive ROI packing algorithm. 
%Further, we can vary the inter-anchor distance $d$ to achieve a speed-accuracy tradeoff. This tradeoff is explored in section \ref{sec:results}.
%Further, there are two parameters we can tune to achieve a speed-accuracy tradeoff. Firstly, an obvious parameter that can be tuned is the inter-anchor distance $d$. Also, most of the CNN object detectors have some amount of robustness inbuilt to tackle scale variations. Hence, we could fit the ROIs extracted from a $s_{1} \times s_{1}$ sized frame into a $s_{2} \times s_{2}$ frame and downscale the ROI packed frame to $s_{3} \times s_{3}$ to get higher speed-ups. There will be some loss of accuracy due to the downscaling. We explore these trade-offs in section \ref{sec:results}.
\begin{figure}
\centering
\begin{tikzpicture}[%
    >=triangle 60,              % Nice arrows; your taste may be different
    start chain=going below,    % General flow is top-to-bottom
    node distance=6mm and 60mm, % Global setup of box spacing
    every join/.style={norm},   % Default linetype for connecting boxes
    ]
\tikzset{
  base/.style={draw, on chain, on grid, align=center, minimum height=4ex},
  proc/.style={base, rectangle, text width=25em},
  test/.style={base, diamond, aspect=2, text width=10em},
  term/.style={proc, align=center, rounded corners},
  % coord node style is used for placing corners of connecting lines
  coord/.style={coordinate, on chain, on grid, node distance=22mm and 60mm},
  % nmark node style is used for coordinate debugging marks
  nmark/.style={draw, cyan, circle, font={\sffamily\bfseries}},
  % -------------------------------------------------
  % Connector line styles for different parts of the diagram
  norm/.style={->, draw},
  free/.style={->, draw, lcfree},
  cong/.style={->, draw, lccong},
  it/.style={font={\small\itshape}}
}

\node [proc, join] (p1) {\small Find all connected components in the ROI graph};
\node [proc, join] (p2) {\small Find the enclosing bounding box around the union of bounding boxes in each connected component and form new set of bounding boxes.};
\node [test, join] (p3) {\small Do the new set of bounding boxes intersect?};
\node [proc] (p4) {\small Determine layout of bounding boxes in the reduced-size image};
\node [proc,join] (p5) {\small Check which pairs of bounding boxes can potentially intersect when expanded in either the horizontal or vertical direction.};
\node [proc,join] (p6) {\small Decide order of dimensions for expansion};
\node [proc,join] (p7) {\small Extend the bounding boxes along each dimension in chosen order.};
\node [proc,join] (p8) {\small Extract patches from the original image according to the final bounding boxes and place them according to the decided layout};
\node [coord, left=of p3, xshift=5em]  (c4)  {};  

\path (p3.south) to node [near start, xshift=1.15em] {$No$} (p4);
  \draw [*->] (p3.south) -- (p4);v 
\path (p3.west) to node [xshift = 0.1em, yshift=-1em] {$Yes$} (c4); 
  \draw [*->] (p3.west) -- (c4) |- (p1); 
\end{tikzpicture}
\caption{ROI packing algorithm. Sample results of the algorithm provided in Figure \ref{fig:illustration}.} \label{fig:flow1}
\end{figure}
\par
\begin{figure}
\begin{tikzpicture}[%
    >=triangle 60,              % Nice arrows; your taste may be different
    start chain=going below,    % General flow is top-to-bottom
    node distance=6mm and 45mm,>=latex]
 \tikzset{
   base1/.style={draw, on chain, on grid, align=center, minimum height=5ex},
   base2/.style={draw, on chain, on grid, align=left, minimum height=5ex},   
   proc1/.style={base1, rectangle, text width=5em},
   proc2/.style={base1, rectangle, text width=20em},
   proc3/.style={base1, rectangle, text width=15em},
   root/.style={base1, rectangle, text width=15em},
   test1/.style={base2, diamond, align=center, text width=5em},
   term1/.style={proc2, align=center, rounded corners}
   }
\node [test1,xshift=0em] (p1) {\small Is the largest dimension height of a bounding box?};
\node [proc1,left= of p1] (p2) {\small Similar set of steps as the yes condition (Symmetrically opposite)};
\node [root,below of = p1,yshift=-20mm] (p3) {\small Decide to extend all bounding boxes horizontally first};
\node [root,below of = p3] (p4) {\small switch(no. of bounding boxes)};
\node[proc1,below of = p4,xshift=-40mm,yshift=-40mm] (c1) {\tiny Single BBOX fills the entire image};
\node[proc1,below of = p4, xshift = -20mm,yshift=-40mm] (c2) {\tiny Place the 2 BBOX's horizontally by their side};
\node[proc1,below of = p4, xshift = 0mm,yshift=-40mm] (c3) {\tiny The BBOX with the largest height is placed in a separate column and the other two BBOXes are placed vertically stacked horizontal to the first BBOX};
\node[proc1,below of = p4, xshift = 20mm,yshift=-40mm] (c4) {\tiny BBOX with the largest height and 3rd largest height are placed in the first column stacked vertically. Other 2 BBOXes are placed in the 2nd column.};
%\path (p0.south) to node [near start, xshift=1em] {} (p1);
%  \draw [->] (p0.south) -- (p1); 
\path (p1.west) to node [near start, xshift=0.3em, yshift=1em] {$No$} (p2);
  \draw [*->] (p1.west) -- (p2); 
\path (p1.south) to node [near start, xshift=1.2em] {$Yes$} (p3);
  \draw [*->] (p1.south) -- (p3);
\path (p3.south) to node [near start, xshift=1em] {} (p4);
  \draw [->] (p3.south) -- (p4);
\path (p4.south) to node [near start, xshift=0.3em] {$1$} (c1);
  \draw [->] (p4.south) -- (c1);
\path (p4.south) to node [near start, xshift=0.3em] {$2$} (c2);
  \draw [->] (p4.south) -- (c2);
\path (p4.south) to node [near start, xshift=0.3em] {$3$} (c3);
  \draw [->] (p4.south) -- (c3);
\path (p4.south) to node [near start, xshift=0.3em] {$4$} (c4);
  \draw [->] (p4.south) -- (c4);
\end{tikzpicture}
\caption{Procedure to determine layout of ROIs in a frame. Sample outputs, including cases with 1, 2 and 4 non-overlapping bounding boxes, are shown in Figure \ref{fig:illustration}.}
  \label{fig:tiling}
\end{figure}
\section{Experimental Methodology}
\label{sec:experiment}
The ImageNet object detection dataset (DET) is a dataset comprising
200 classes of objects that form a subset of the ImageNet 1000
classes. Further, the ImageNet video object detection dataset (VID)
comprises of 30 classes of objects from among the DET 200 classes. The
ImageNet video object detection (VID) dataset was the most appropriate
choice for illustrating the results of our work. The ImageNet VID
training set has 3862 video snippets and the ImageNet VID validation
set has 555 video snippets. 53539 frames from the DET dataset
comprising only of the classes from the VID dataset and 57834 frames
from the VID training set were combined to form the final training set
in our experiments.
%This information is summarized in figure \ref{tab:dataset}. 
\par 
The SSD300 \cite{liu2016ssd} object detector operated on a per-frame basis was used as the baseline for our work. The SSD300 object detector uses VGG16 as feature extractor. The SSD300 pretrained model on the DET dataset was further trained on our training set for 210k iterations with a learning rate of $10^{-3}$ for the first $80000$ iterations, $10^{-4}$ for the next $40000$ iterations and $10^{-5}$ for the rest of the training. This SSD300 trained model gave a mAP score of $70.6$ on the VID validation set. Further, this model has a network throughput of $47$ fps and a overall throughput (including standard pre-processing time) of $18$ fps.
%\par
%\begin{figure}
%\begin{centering}
%\begin{tabular}{|p {2cm}|p {1cm} |p {1cm} |p {1cm} |}
%\hline 
%Dataset & No. of classes & No. of video snippets & No. of frames selected\tabularnewline
%\hline 
%\hline 
%DET training set & 200 & N/A & 53539\tabularnewline
%\hline 
%VID training set & 30 & 3862 & 57834\tabularnewline
%\hline 
%VID validation set & 30 & 555 & 176126\tabularnewline
%\hline 
%\end{tabular}\caption{Details of the ImageNet DET and ImageNet VID dataset}\label{tab:dataset}
%\par\end{centering}
%\end{figure}
The SSD300 network processes images at $300\times300$ as the name
suggests. However, closer observation of the network suggested that
the same network can process $150\times150$ images as well by stopping
processing at the penultimate layer. Hence, we use the same SSD300
network to process both full-size and reduced-size images. In all our
experiments, the full size $s_{1}$ is $300$ and the reduced size
$s_{2}$ is $150$. When a $150\times150$ sized image is passed on to
the SSD300 network, processing is configured to stop at the
penultimate layer. All the experiments were performed using the SSD
Caffe framework running on a 2.1 GHz Intel Xeon CPU with a Nvidia
TITAN X GPU. Code will be released soon. In all the experiments, the
batch size was $1$ to emulate a real-time processing scenario. The
detection threshold $\tau$ used to select ROIs was fixed at $0.2$ in
our experiments unless explicitly specified otherwise.
\section{Experimental Results}
\label{sec:results}
\subsection*{Results from sample videos}
\begin{figure}
\centering
%\advance\leftskip-1cm
%\advance\rightskip-3cm
\begin{tabular}{c|c|c}
Frame $i$ & ROI-packed frame $i+1$ & Frame $i+1$\\
\hline
\subfloat{\includegraphics[scale=0.08]{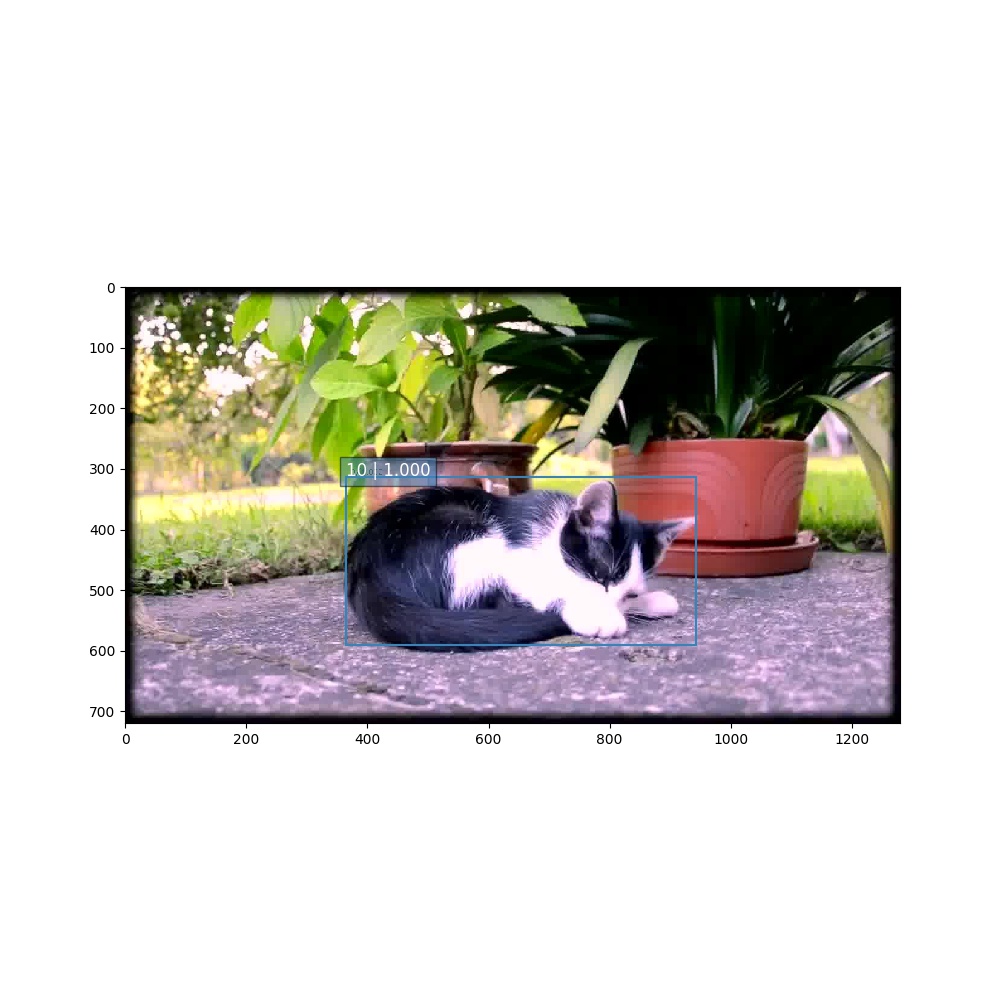}} &
\subfloat{\includegraphics[scale=0.08]{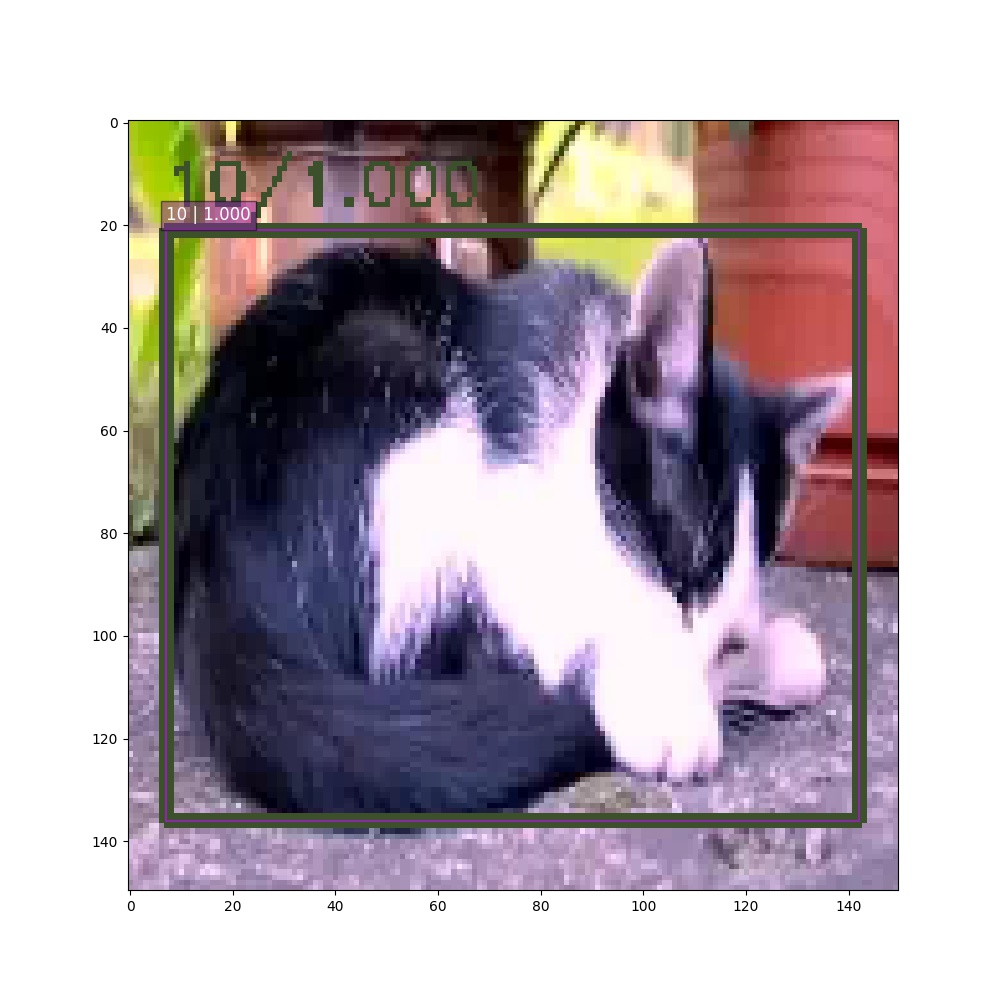}} &
\subfloat{\includegraphics[scale=0.08]{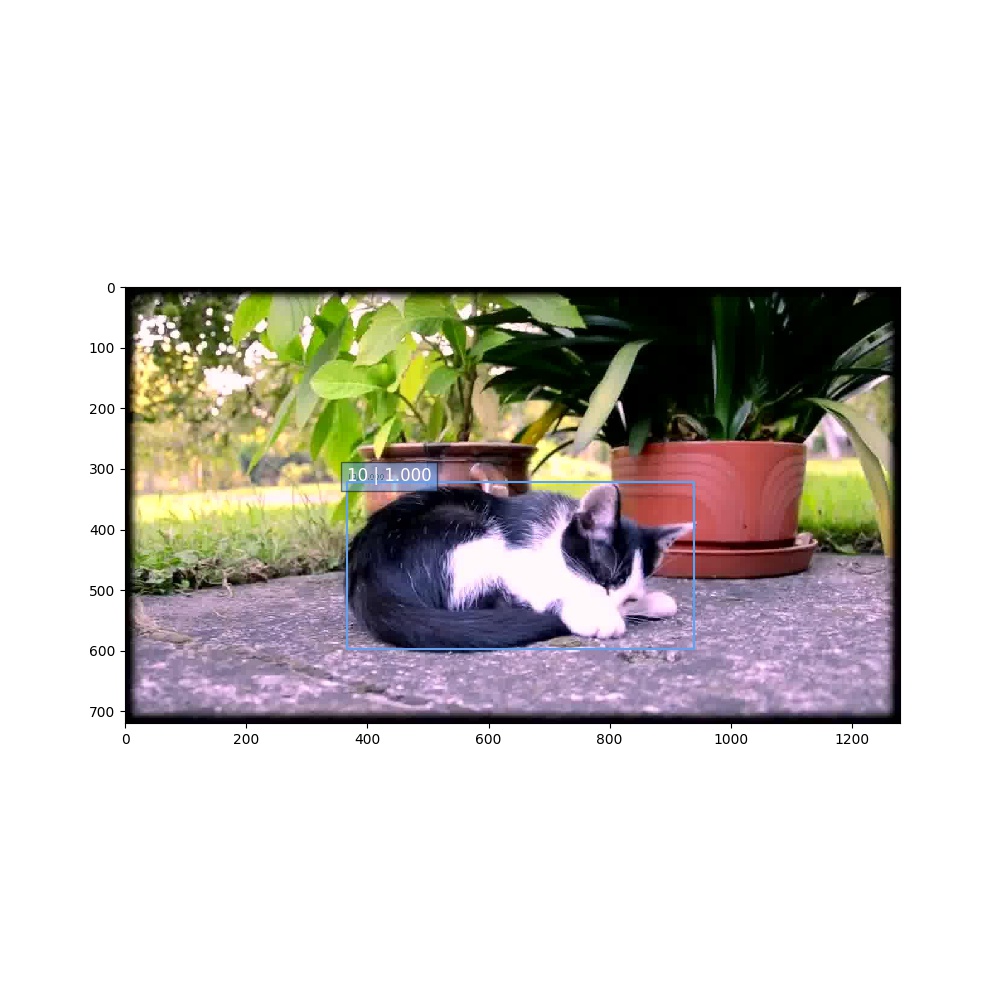}} \\[-0.3in]

\subfloat{\includegraphics[scale=0.08]{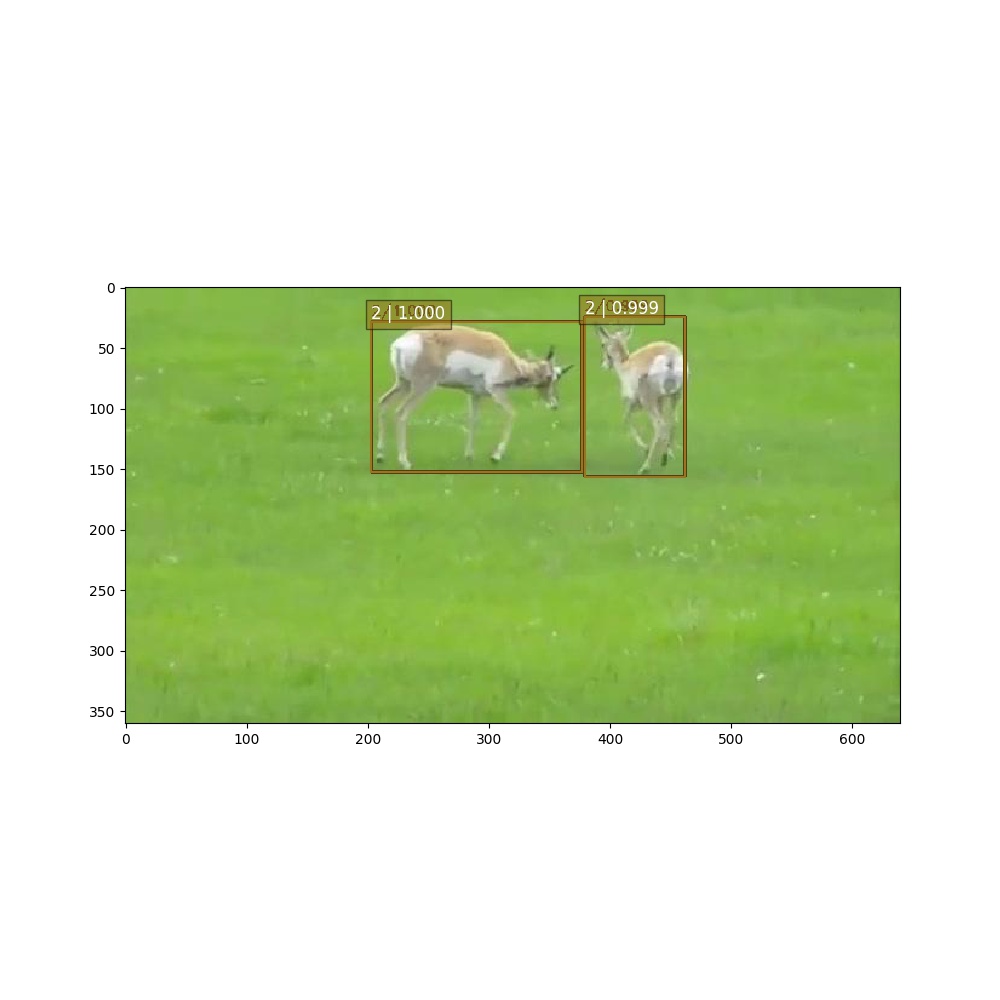}} &
\subfloat{\includegraphics[scale=0.08]{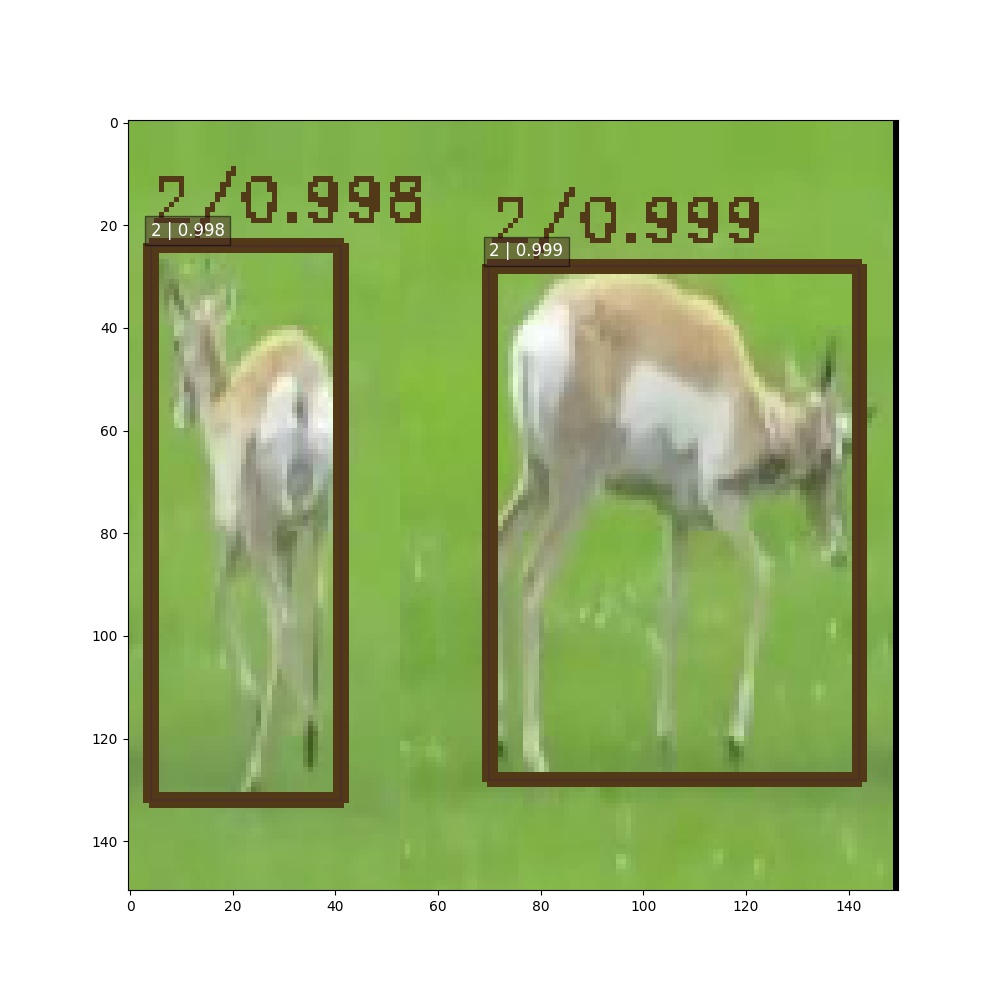}} &
\subfloat{\includegraphics[scale=0.08]{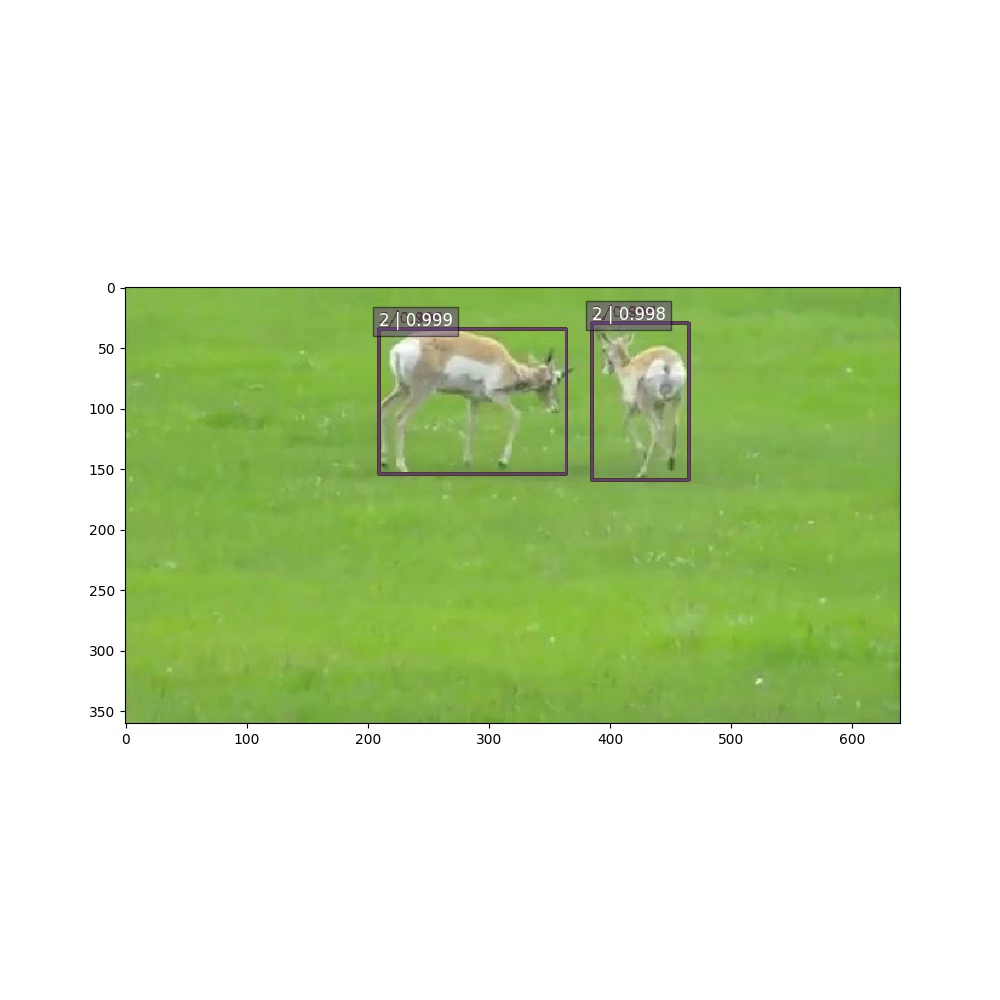}} \\[-0.3in]

\subfloat{\includegraphics[scale=0.08]{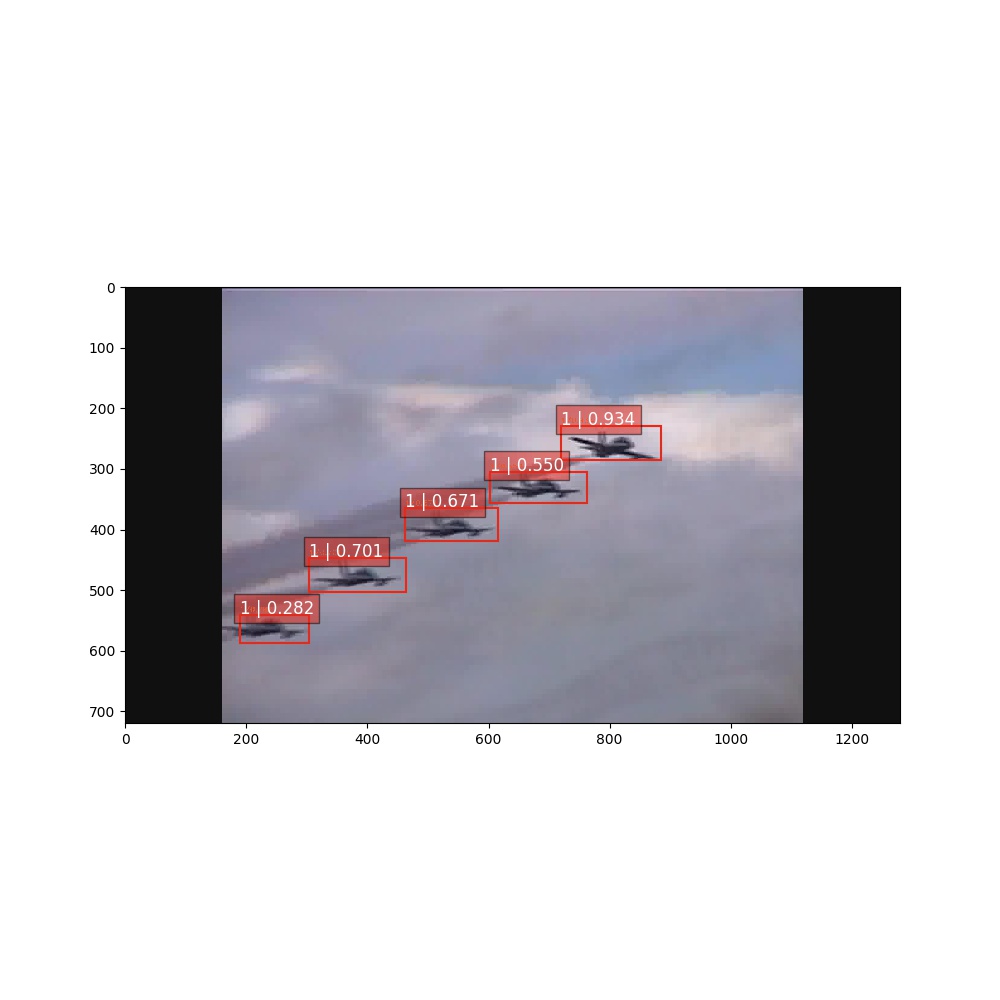}} &
\subfloat{\includegraphics[scale=0.08]{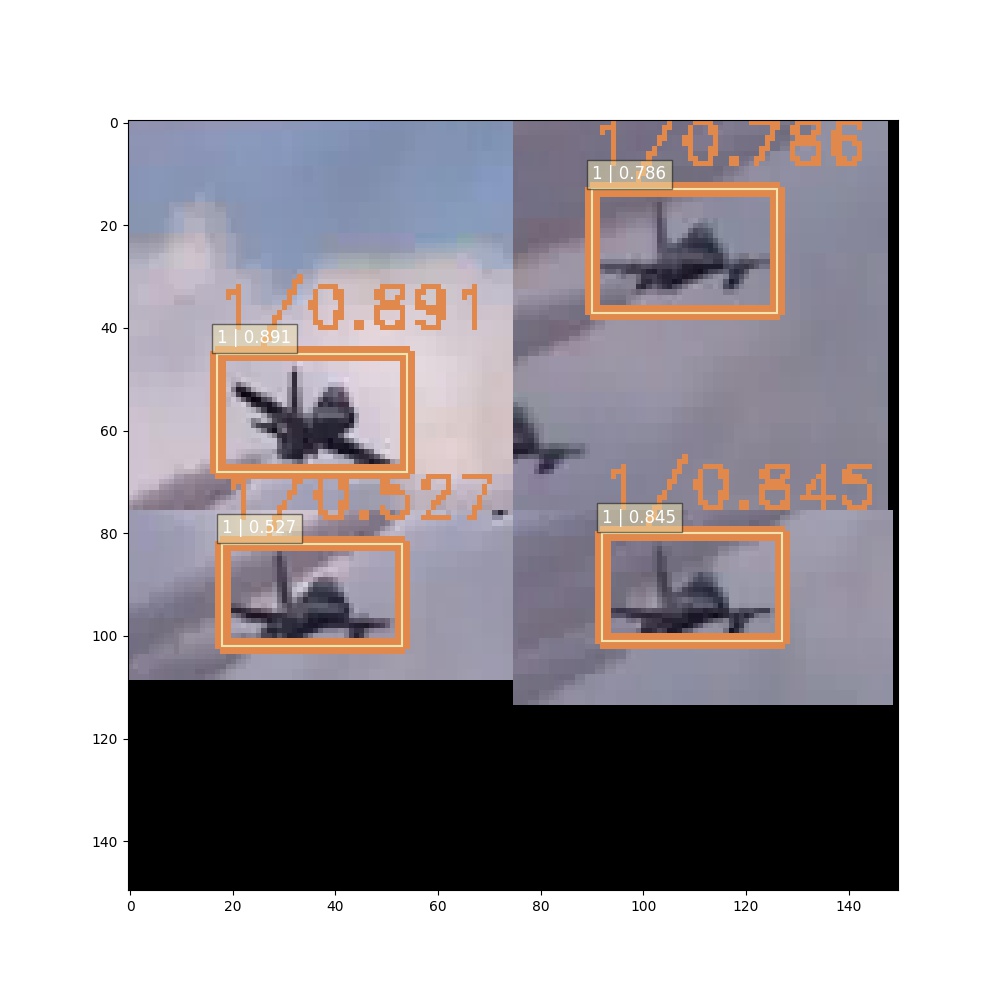}} &
\subfloat{\includegraphics[scale=0.08]{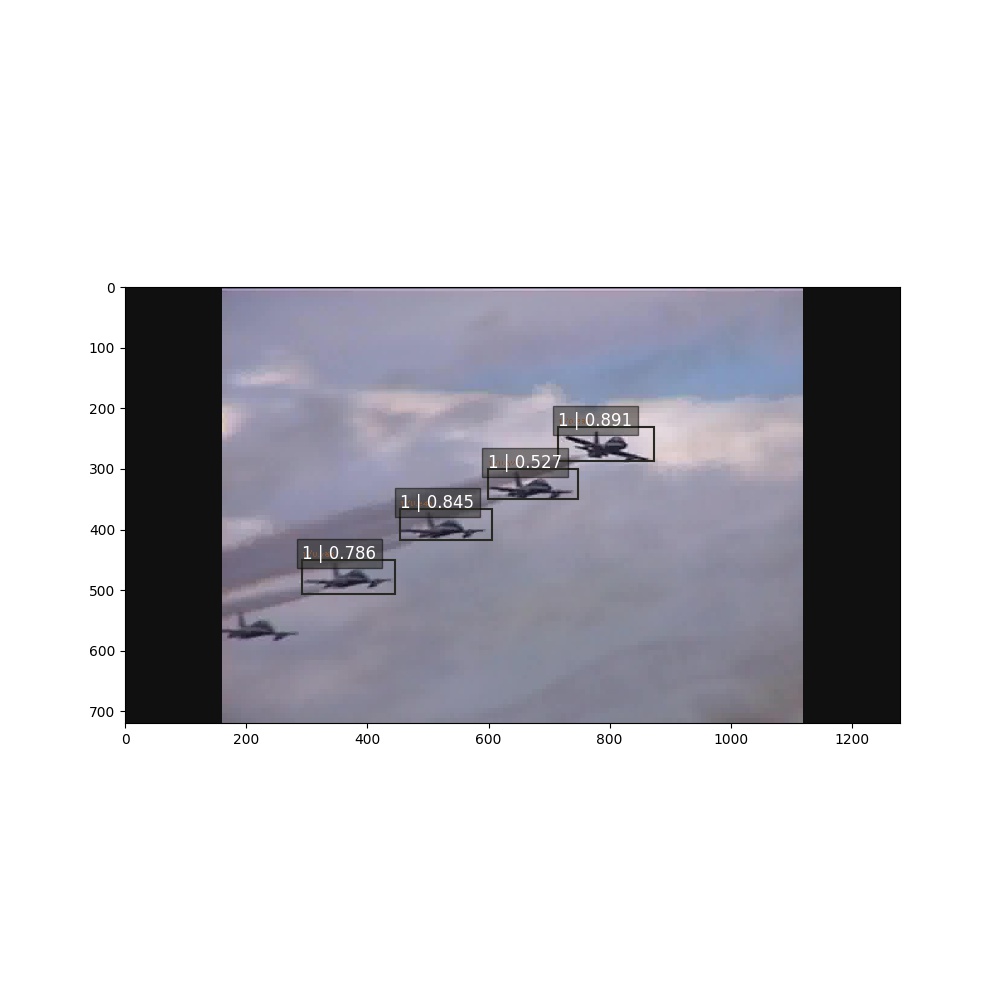}} \\[-0.3in]

\subfloat{\includegraphics[scale=0.08]{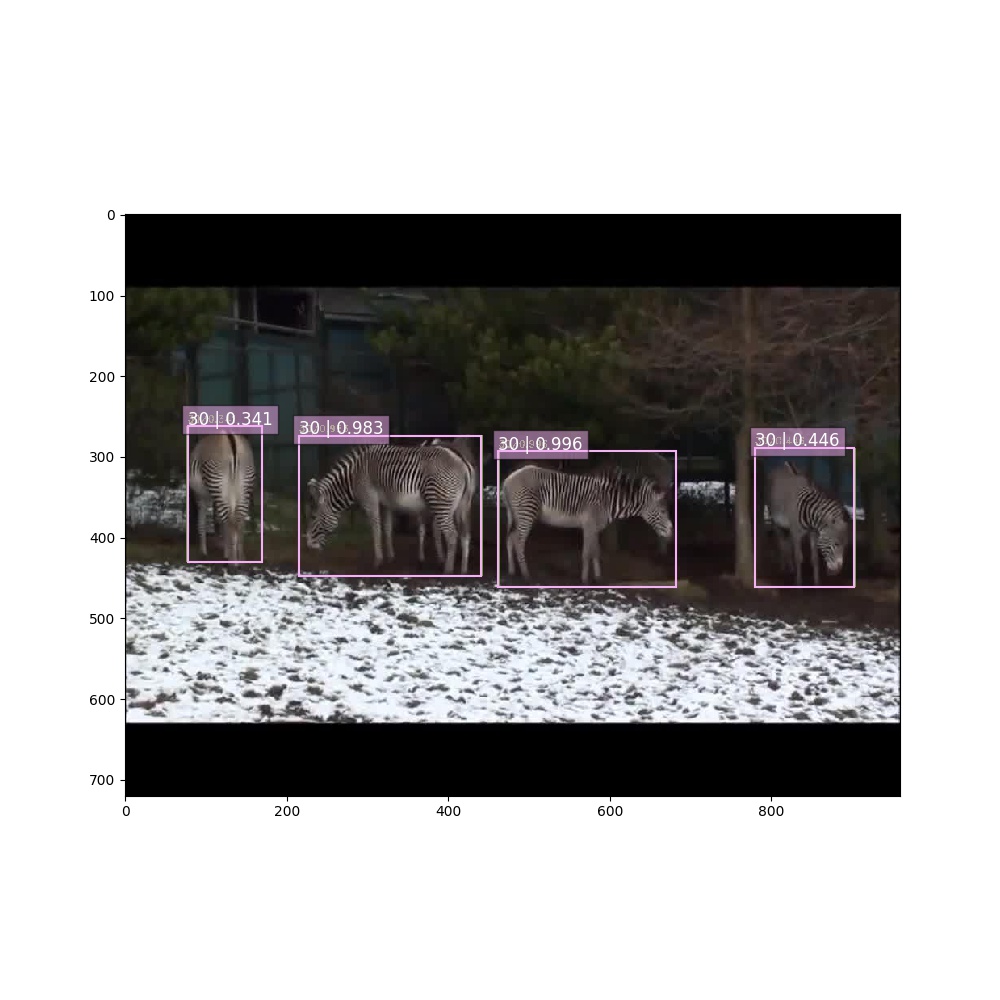}} &
\subfloat{\includegraphics[scale=0.08]{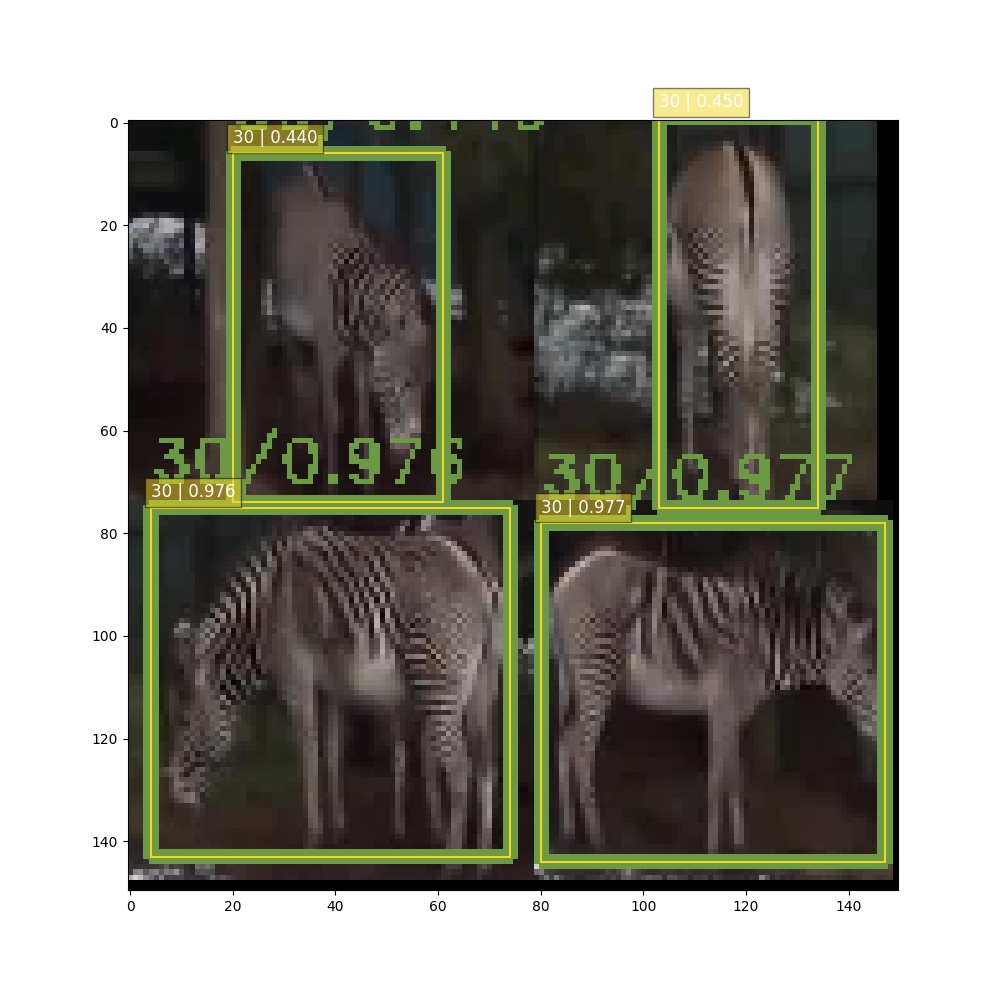}} &
\subfloat{\includegraphics[scale=0.08]{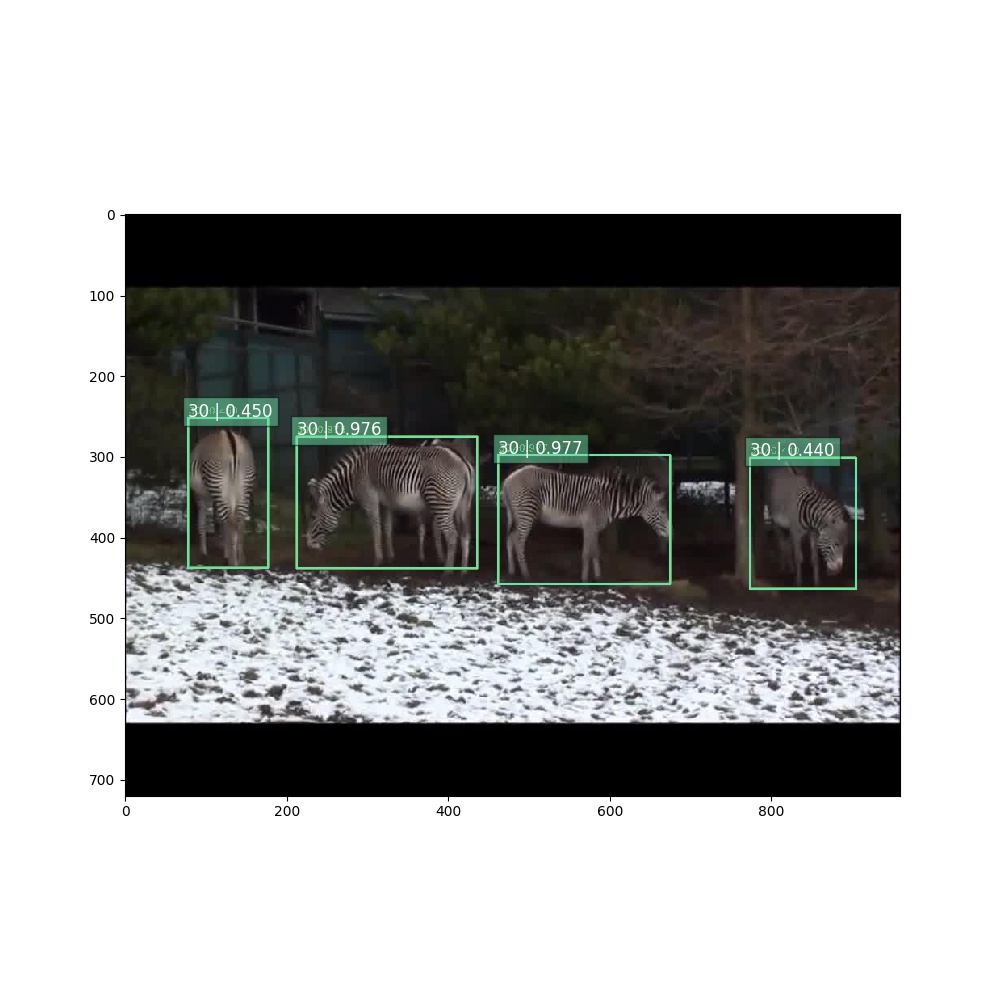}} \\

\end{tabular}
\caption{Consecutive frames processed with the ROI packing algorithm. The first column shows frame $i$. The second column shows the ROI packed frame $i+1$ with the detections. The third column shows the original frame $i+1$ with detections transformed from ROI packed frame $i+1$.}
\label{fig:illustration}
\end{figure}

We show results on processing some sample videos with PaD. Figure
\ref{fig:illustration} provides sample detections with our ROI-packing
algorithm. The first column shows frame $i$. The second column shows
the ROI-packed reduced-size frame $i+1$ with the detections. The third
column shows the original frame $i+1$ with detections mapped from
ROI-packed frame $i+1$. For this experiment, the confidence threshold
$\tau$ for selecting a detection as an ROI for the next frame was set
to $0.3$ for the sake of illustration. All bounding boxes with a
minimum threshold of $0.2$ are shown in the figure.
\par
\begin{figure}
  \centering
  \begin{tabular}{c}
  \subfloat[]{\includegraphics[scale=0.45]{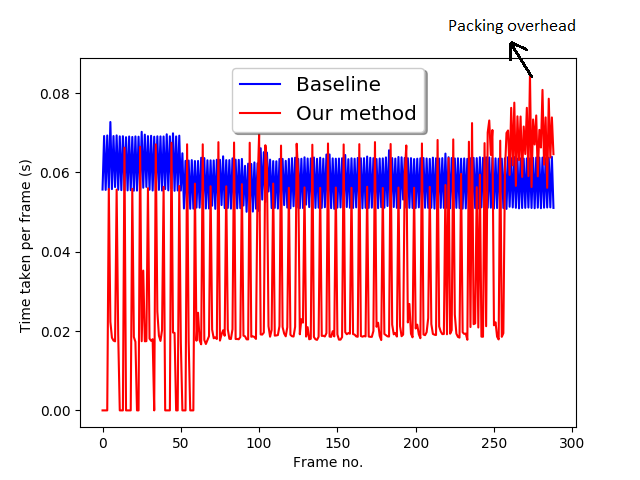}} \\
\subfloat[]{\includegraphics[scale=0.38]{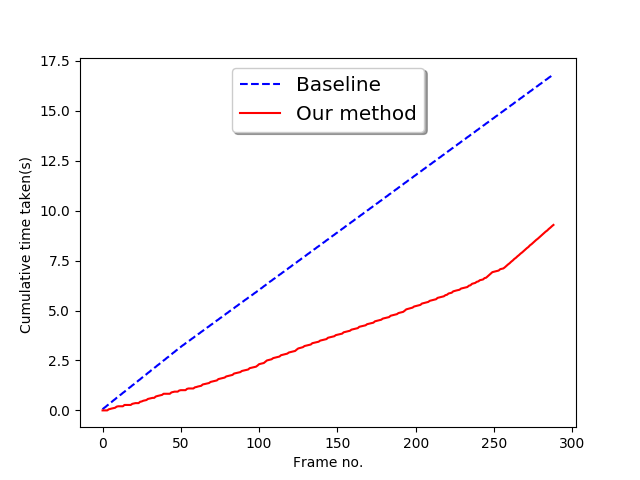}} \\
  \end{tabular}
  \caption{Comparing speed of PaD to the baseline for a sample video:
    (a) Per-frame processing time and (b) Cumulative processing time}
  \label{fig:tt}
\end{figure}

 In Figure \ref{fig:tt}, we plot the per-frame time as well as the
cumulative time for processing a sample video using PaD and the
baseline. It can be seen that processing the lower sized frame of
$150\times150$ is almost $3\times$ faster. When ROI packing fails,
there is a slight overhead incurred which is visible towards the end
of the video in Figure \ref{fig:tt}(a). Also, we see that some frames
require almost no time for processing. This is because the previous
frame had no detections. Overall, from Figure~\ref{fig:tt}(b), we note
that processing the video using PaD requires almost $8$s lesser time
than the baseline.

\subsection*{Results over the entire dataset}
\begin{figure}[H]
  \centering
  \includegraphics[scale=0.5]{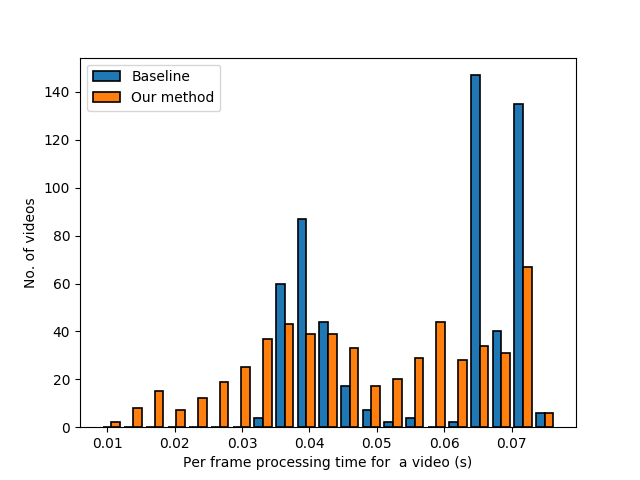}
  \caption{Histogram of average per-frame processing time on a video-by-video basis}
  \label{fig:tpvpf}
\end{figure}

PaD was run with a inter-anchor distance $d=5$ and $s_{2}=150$. In
Figure \ref{fig:tpvpf}, we plot the histogram of average per-frame
processing time on a video-by-video basis. In other words, the average
time taken per frame was obtained for each video and is plotted as a
histogram across videos. From the figure, we can clearly see that the
average time taken to process a frame is lower using PaD for more
videos than the baseline. The average per-frame speedup is around
$1.25 \times$ and the FLOP reduction on the average is $32\%$. The
average overhead incurred for ROI-packing is around $9\%$ of the total
time taken.  The mAP score drops by $1.1\%$ (from $70.6$ to $69.5$).

\subsection*{Comparison with a naive ROI-packing algorithm}
In order to illustrate the benefits of our ROI-packing algorithm
discussed in section \ref{sec:Methodology}, we compare the accuracy
drop when compared with a naive ROI-packing algorithm.

\par The naive ROI-packing algorithm can accommodate upto four ROIs
just like the sophisticated method. If there are more than four
objects in the frame, the frame is processed at full size. Otherwise,
the bounding box surrounding each frame is extended by a factor of
$1.2\times$ and is treated as an ROI. If there is only one object, the
ROI surrounding the bounding box is rescaled to size $s_{2} \times
s_{2}$ and is processed by the detector. If there are two objects, the
lower sized frame is divided into two columns of size $s_{2} \times
\frac{s_{2}}{2}$. The two ROIs are rescaled to the appropriate sizes
and laid out on the lower sized frame. In the case of three or four
objects, the lower sized frame is divided into four regions in two
columns and two rows of size $\frac{s_{2}}{2} \times
\frac{s_{2}}{2}$. In the case of three ROIs, the ROIs will be rescaled
to occupy three of the four regions in the frame and the fourth region
will be left blank. In the case four ROIs, the ROIs will be rescaled
and fit to these four regions. We do not perform a greedy expansion of
the RoIs to provide additional background context. Instead, the ROIs are just
expanded by a constant factor of $1.2 \times$ and rescaled to
appropriate size.

\par PaD's ROI packing method with inter-anchor distance $d=5$ gave a
mAP score of $69.5$. With the same parameter setting, the naive ROI
packing algorithm gave a mAP score of $56.8$. This clearly illustrates
the need for an ROI-packing algorithm that preserves the
scale and aspect ratio of the ROIs and provides as much background
context as possible.
\section{Conclusion and Future Work}
\label{sec:conclusion}
Still-image object detection has improved by leaps and bounds in
recent years due to the success in training and deploying neural
networks. However, the opportunities that are available in the context
of videos have not been fully exploited. Neural networks are in
general very compute-intensive. In this work, we use the opportunities
available in the context of videos to speed up and reduce the amount
of computation in neural network based object detectors. In the
proposed method, called PaD, the full-sized input is only processed in
selected anchor frames. In the inter-anchor frames, ROIs are
identified based on the locations of objects in the previous
frame. These ROIs are packed together in a reduced-size frame that is
fed to the CNN object detector. The ROI packing algorithm needs to
ensure that the scales and aspect ratios of the objects are preserved
and enough background context is provided. With this setup, we
observed $1.25 \times$ speedup with $1.1\%$ drop in accuracy
on the ImageNet VID validation set. Further, the time taken to process
a lower sized frame is almost $3 \times$ lesser and the FLOP count
reduces by $4 \times$.
%Given more suitable datasets, we can get even
%more speedup and reduction in FLOP count in the average sense.
\par
As part of future work, we plan to incorporate a motion model to
obtain the ROIs in the current frame. Incorporating a motion model
could also help extend this framework to larger batch sizes. Also, it
is possible to use two different models or networks to process larger
sized and smaller sized frames. This will help reduce the accuracy
drop but will in turn increase the memory footprint. There is an
overhead incurred in checking whether the ROIs can fit in the lower
sized frame. Currently, we select anchor frames at regular
intervals. However, information on whether ROIs were packed
successfully in previous frames can help us decide how frequently we
select anchor frames. Thus, another line of future work is a dynamic
mechanism for selecting anchor frames in order to reduce the
overhead. It would be interesting to test PaD in more resource
constrained platforms like mobile GPUs and CPUs. We expect the
benefits to be more pronounced in such platforms.
%Further, instead of using a heuristic hand-crafted algorithm for forming the ROI packed frame, a neural network could be trained to identify the ROIs and pack them in a lower sized frame like in an attention mechanism. This is another ambitious line of future work. It would be interesting to test PaD in more resource constrained platforms like mobile GPUs and CPUs. We expect the benefits to be more pronounced in such platforms.  
\begin{acks}
This work was supported by Intel India, the Robert Bosch Centre for
Data Science and AI (RBC-DSAI) and the Center for Computational Brain
Research (CCBR) at IIT Madras.
\end{acks}